\documentclass[a4paper, 10pt, conference]{ieeeconf}      % Use this line for a4
\usepackage{FG2024}
\usepackage{cite}
\usepackage{hyperref}
\usepackage{amsmath,amssymb,amsfonts}
\usepackage{algorithmic}
\usepackage{graphicx}
\FGfinalcopy % *** Uncomment this line for the final submission

\IEEEoverridecommandlockouts                              % This command is only
\def\FGPaperID{125} % *** Enter the FG2024 Paper ID here

\title{\LARGE \bf
Audio-Visual Person Verification based on Recursive Fusion of \\ Joint Cross-Attention
}

\author{\parbox{16cm}{\centering
    {\large R. Gnana Praveen and Jahangir Alam}\\
    {\normalsize
    Computer Research Institute of Montreal (CRIM), Canada}}
    \thanks{The authors wish to acknowledge the funding from the Government of Canada’s New Frontiers in Research Fund (NFRF) through grant NFRFR-2021-00338.
    }% <-this % stops a space
}

\begin{document}

\ifFGfinal
\thispagestyle{empty}
\pagestyle{empty}
\else
\author{Anonymous FG2024 submission\\ Paper ID \FGPaperID \\}
\pagestyle{plain}
\fi
\maketitle

%%%%%%%%%%%%%%%%%%%%%%%%%%%%%%%%%%%%%%%%%%%%%%%%%%%%%%%%%%%%%%%%%%%%%%%%%%%%%%%%
\begin{abstract}
Person or identity verification 
%Conventionally, person or identity verification has been explored using audio modality in the literature and achieved remarkable success using state-of-the-art deep learning models. However, exploring audio and visual modalities 
has been recently gaining a lot of attention using audio-visual fusion as faces and voices share close associations with each other. Conventional approaches based on audio-visual fusion rely on score-level or early feature-level fusion techniques. Though existing approaches showed improvement over unimodal systems, the potential of audio-visual fusion for person verification is not fully exploited. %for the task of speaker verification as they often exhibit complementary relationships.  
In this paper, we have investigated the prospect of effectively capturing both intra- and inter-modal relationships across audio and visual modalities, which can play a crucial role in significantly improving the fusion performance over unimodal systems. In particular, we introduce a recursive fusion of a joint cross-attentional model, where a joint audio-visual feature representation is employed in the cross-attention framework in a recursive fashion to progressively refine the feature representations that can efficiently capture the intra- and inter-modal relationships. To further enhance the audio-visual feature representations, we have also explored BLSTMs to improve the temporal modeling of %intra-modal relationships pertinent to 
audio-visual feature representations. Extensive experiments are conducted on the Voxceleb1 dataset to evaluate the proposed model. Results indicate that the proposed model shows promising improvement in fusion performance by adeptly capturing the intra- and inter-modal relationships across audio and visual modalities. Code is available at 
\url{https://github.com/praveena2j/RJCAforSpeakerVerification}    
\end{abstract}

%%%%%%%%%%%%%%%%%%%%%%%%%%%%%%%%%%%%%%%%%%%%%%%%%%%%%%%%%%%%%%%%%%%%%%%%%%%%%%%%
\section{INTRODUCTION}

%%% Speaker verification
Person verification deals with the problem of verifying the identity of a person, which has a wide range of applications in various fields such as forensics, commercial, and law enforcement applications \cite{7298570}. The task of person verification has been predominantly explored using faces \cite{WANG2021215} and speech \cite{9889705} signals independently. With the advancement of deep learning models, both face- and speech-based methods   
%using speech as the primary modality in the literature. By leveraging the advancement in deep learning models, speech-based methods 
have individually shown significant improvement in the performance of person verification \cite{9889705}. However, relying on individual modalities may often deteriorate the performance of the system when face or speech-based signals are degraded by extreme background noise or intra-variations such as pose, low illumination, manner of speaking, etc. Therefore, leveraging the fusion of both faces and voices has been gaining momentum as 
%speech can be interrupted by extreme background noise or other speakers in real-world scenarios. Additionally, speech signals are also prone to a large degree of variability due to age, manner of speaking, etc. Therefore, multimodal learning has been gaining a lot of attention as 
multiple modalities are often expected to complement each other \cite{10096814}. For instance, when speech modality is corrupted, we can rely on other modalities such as face to verify the identity of a person and vice-versa. Effectively leveraging both inter-modal complementary associations and intra-modal relationships among the audio and visual modalities plays a crucial role in significantly outperforming unimodal approaches \cite{10095234}. %The close association between faces and voices has also been investigated for cross-modal verification tasks by learning a common representation for both modalities \cite{}. %Though person verification can also be achieved using modalities such as iris, fingerprints, gait, etc., face and voice are the widely explored contact-free channels as they are closely associated with each other \cite{8578977}.   
Inspired by the close association between faces and voices, several approaches have been explored for cross-modal verification tasks by learning a common representation for both modalities \cite{Nagrani18a,Nagrani18c}. Most of the existing approaches for person verification based on audio-visual (A-V) fusion focused on score-level fusion \cite{251181,FITPUB12292} or early feature-level fusion \cite{Chen2020,9350195}. Score-level fusion refers to the fusion of scores obtained from vocal and facial embeddings of the individual modalities, whereas early feature-level fusion refers to the aggregation of the embeddings of audio and visual modalities using simple feature concatenation or averaging. Though these methods have improved the fusion performance over unimodal systems, they fail to leverage the rich complementary inter-modal relationships across the audio and visual modalities. %Ideally, an efficient A-V fusion should capture the complementary inter-modal relationships, while retaining the benefits of intra-modal relationships.   

%Though faces and voices have been independently studied for the task of identifying the person, the potential of audio-visual fusion for speaker verification is not fully exploited in the literature. 
%Traditionally, multimodal fusion can be categorized into three major strategies: decision-level fusion, feature-level fusion, and hybrid-level fusion \cite{wu_lin_wei_2014}. In decision-level fusion (late fusion), multiple modalities are trained independently, and their individual predictions are fused to produce the final predictions. While decision-level fusion is straightforward to implement and requires less training, it overlooks interactions among the individual modalities, resulting in limited improvement over unimodal approaches. Conversely, feature-level fusion (early fusion) involves concatenating features from audio and visual modalities immediately after extraction and using them for final predictions. Hybrid fusion combines the advantages of both decision-level and feature-level fusion by integrating outputs from both approaches. Most of the current approaches rely on decision-level fusion \cite{10096814} or simple feature concatenation \cite{8683477,8683477}. In both of these categories, they fail to exploit the rich inter-modal relationships across the audio and visual modalities. 

In recent years, attention-based models have been explored to efficiently capture the complementary inter-modal associations across faces and voices \cite{sun23_interspeech,9320237}. Most of the existing attention-based models attempted to leverage the intra- and inter-modal relationships in a decoupled fashion \cite{10095883}. %either focus on intra-modal relationships or inter-modal relationships but not both. 
Another line of approaches focused on dealing with noisy modalities using a weighted combination of audio and visual modalities \cite{10096814,8683477}. To effectively fuse audio and visual modalities, it is important to adeptly capture both intra-modal relationships (temporal dynamics of videos) and inter-modal relationships (complementarity of the modalities). %Intra-modal relationships offer rich information pertinent to the temporal dynamics of videos, whereas inter-modal relationships provide significant information related to the complementarity of the modalities. 
Contrary to the prior approaches, we have explored joint cross-attentional fusion in a recursive fashion to simultaneously enhance the modeling of both intra- and inter-modal relationships to obtain robust A-V feature representations. Recursive attention has been previously explored successfully for emotion recognition \cite{10095234} and event localization \cite{9423042}.
%Therefore, we have explored the prospect of effectively capturing both intra- and inter-modal relationships using the recursive fusion of audio and visual features obtained from the joint cross-attentional model. 
By recursively fusing the features of audio and visual modalities we can achieve more refined feature representations progressively in order to improve the performance of A-V fusion for person verification. The major contributions of the proposed approach can be summarized as follows: (1) A recursive fusion of joint cross-attentional model is introduced to efficiently capture both intra- and inter-modal relationships across faces and voices for person verification. (2) Bidirectional Long Short-Term Memory (BLSTM) networks are explored to further improve the modeling of temporal dynamics %intra-modal relationships (temporal dynamics) of 
pertinent to A-V feature representations. %To improve the intra-modeling (temporal dynamics) of A-V feature representations, we have further integrated transformers with the recursive fusion. 
(3) Extensive experiments are conducted on the Voxceleb1 dataset to evaluate the robustness of the proposed model. %and outperform the state-of-the-art A-V fusion models. 

%%%% Current methods on A_V fusion

%%%% Motivation

%%%% Summary of contributions

%\begin{itemize}
%    \item A recursive fusion of joint cross-attentional model is introduced to efficiently capture both intra- and inter-modal relationships across faces and voices for person verification.
%    \item Bidirectional Long Short-Term Memory (BLSTM) networks are further explored to improve the modeling of intra-modal relationships (temporal dynamics) of A-V feature representations.   
%    \item Extensive experiments are conducted on the voxceleb1 dataset to evaluate the robustness of the proposed model and outperform the state-of-the-art A-V fusion models. 
%\end{itemize}

\section{Related Work}

%The close association between faces and voices has attracted much attention for the task of the cross-modal biometric matching system by projecting the features of individual modalities to a common representation space \cite{Nagrani18a,Nagrani18c}. 
Sari et al. \cite{9414260} explored a multiview approach by transforming the individual feature representations into a common representation space and a shared classifier is used for both modalities for person verification. Chen et al. \cite{10096925} leveraged the complementary information as a means of supervision to obtain robust A-V feature representations using a co-meta learning paradigm in a self-supervised learning framework. Tao et al. \cite{10096814} also explored the complementary relationships between audio and visual modalities to clean noisy samples, where consistency among audio and visual modalities is used to discriminate between easy and hard samples. Another line of approaches is to deal with mitigating the impact of noisy modalities by leveraging complementary relationships. Shon et al. \cite{8683477} proposed an attention mechanism to assign higher attention scores to the modality exhibiting higher discrimination. They leveraged the complementary nature across audio and visual modalities to mitigate the influence of noisy modalities. Hormann et al. \cite{9320237} further extended the idea of \cite{8683477} by introducing feature fusion of audio and visual modalities at intermediate layers to improve the quality of feature representations. Chen et al. \cite{Chen2020} explored the prospect of obtaining robust feature representations by investigating various fusion strategies at the embedding level and achieved the best performance using gating-based fusion. They further exploited data augmentation strategy to deal with extremely corrupted or missing modalities. 

All the above-mentioned approaches fail to leverage the cross-modal interactions to effectively capture the rich inter-modal relationships. Cross Attention (CA) has been successfully explored in several applications such as weakly supervised action localization \cite{lee2021crossattentional}, event localization \cite{9423042}, and emotion recognition \cite{9667055,10005783}. Recently, Mocanu et al. \cite{9922810} explored CA based on cross-correlation across the audio and visual modalities to effectively capture the complementary inter-modal relationships for person verification. Liu et al. \cite{10095883} explored CA by deploying cross-modal boosters in a pseudo-siamese structure to model one modality by exploiting the knowledge from another modality. However, they focus only on inter-modal relationships \cite{9922810} or capture the intra- and inter-modal relationships in a decoupled fashion \cite{10095883}. Unlike these approaches, Praveen et al. \cite{praveen2023audiovisual} introduced joint feature representation in the CA framework and improved fusion performance. % by integrating the joint feature representation in the cross-attention framework. 
In this work, we explored CA in a joint recursive fashion to simultaneously enhance the modeling of intra- and inter-modal relationships. Although \cite{10095234} is closely related to our approach, the proposed approach differs primarily from \cite{10095234} in two aspects: (1) In \cite{10095234}, RJCA is proposed for emotion recognition in the context of continuous regression, whereas we have explored it in the framework of binary classification using AAMSoftmax \cite{8953658} loss for person verification. (2) In \cite{10095234}, frame-level predictions are obtained, while in this work we have used Attentive Statistics Pooling \cite{okabe18_interspeech} to effectively aggregate features across temporal dimensions to obtain robust A-V embeddings at the utterance level, achieving state-of-the-art results.

%The work of Praveen et al \cite{10095234} is close to our approach, however the proposed approach primarily differs from \cite{10095234} in two aspects: (1) In \cite{10095234}, RJCA is proposed for emotion recognition in the context of continuous regression, whereas we have explored it in the framework of binary classification using AAMSoftmax loss for person verification. (2) In \cite{10095234}, frame-level predictions are obtained, whereas in this work, we have employed ASP to effectively aggregate the features across temporal dimensions to obtain robust utterance-level A-V embeddings, achieving state-of-the-art results.   
%Unlike prior approaches, we have explored a recursive fusion of joint cross-attention to effectively capture both intra- and inter-modal relationships to build an efficient A-V system for speaker verification. %Recursive fusion of cross-modal attention models has also been successfully explored for event localization \cite{9423042}, emotion recognition \cite{10095234}, etc. 

%Praveen et al \cite{10005783} explored a joint cross-attentional (JCA) framework for dimensional emotion recognition, which is closely related to our work. However, we have enhanced the JCA model by introducing LSTMs to improve the intra-modeling relationships of the individual modalities as well as the attended feature representations. We have further adapted the JCA modal for speaker verification by introducing the attentive statistics pooling module.   

\section{Methodology}
%\subsection{Problem Formulation}
\noindent \textbf{A) Problem Statement:}
Given an input video subsequence $S$, we uniformly sample $L$ non-overlapping video segments and extract deep feature vectors from pre-trained models for audio and visual modalities. Let ${\boldsymbol X}_{a}$ and ${\boldsymbol X}_{v}$ denote the deep feature vectors of audio and visual modalities respectively for the given input video sub-sequence $S$ of fixed size, which is expressed as 
%\begin{equation}
${ \boldsymbol X}_{a}=  \{ \boldsymbol x_{a}^1, \boldsymbol x_{a}^2,..,\boldsymbol x_{a}^L \} \in \mathbb{R}^{d_a\times L}$ and 
%\end{equation}
%\begin{equation}
${ \boldsymbol X}_{v}=  \{ \boldsymbol x_{v}^1, \boldsymbol x_{v}^2,..,\boldsymbol x_{v}^L \} \in \mathbb{R}^{d_v\times L}$,
%\end{equation}
where ${d_a}$ and ${d_v}$ represent the dimensions of the audio and visual feature vectors, respectively. $\boldsymbol x_{a}^{ l}$ and $\boldsymbol x_{v}^{ l}$ denote the audio and visual feature vectors of the video segments, respectively, for the $l = 1, 2, .., L$ segments. The objective of the problem is to estimate the person verification model $f:\boldsymbol{X} \to \boldsymbol{Y}$ from the training data $\boldsymbol X$, where $\boldsymbol X$ denotes the set of audio and visual feature vectors of the input video segments and $\boldsymbol Y$ represents the person identity of the corresponding subsequence $S$.
 \begin{figure*}[t!]
\centering
\includegraphics[width=0.9 \linewidth]{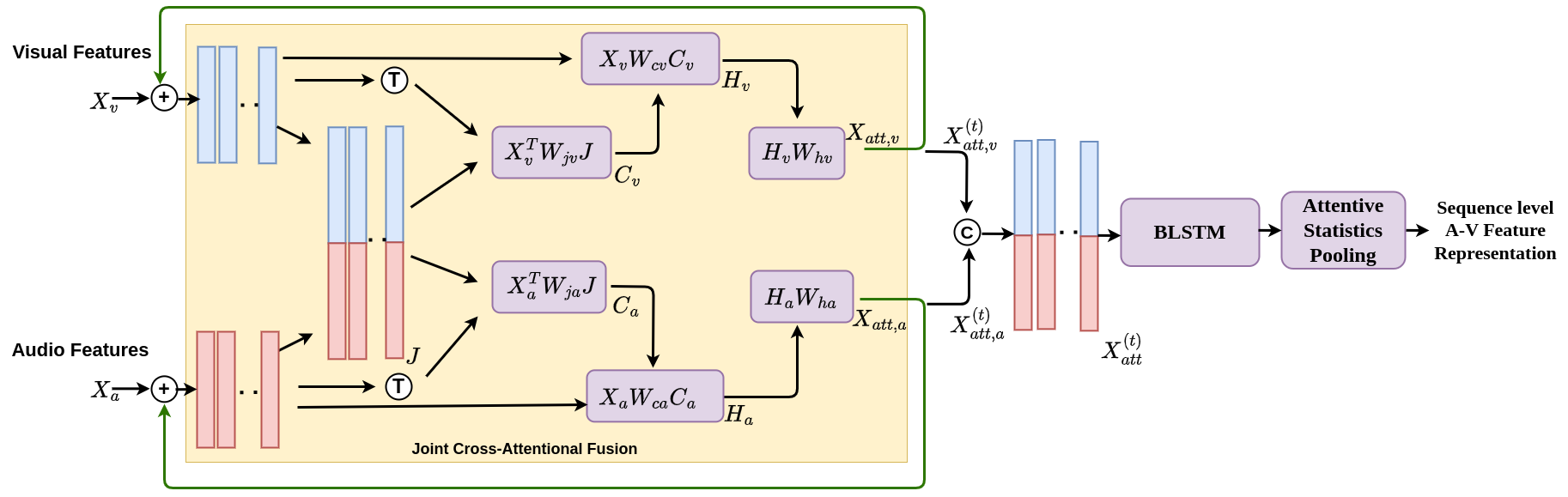}
\caption{\textbf{Block Diagram of the Recursive Joint Cross-Attention model for A-V fusion }}
\label{Block Diagram}
\end{figure*}
% \section{Recursive Joint Cross-Attention}

%\subsection{Recursive Joint Cross-Attention:}
\noindent \textbf{B) Recursive Joint Cross-Attention (RJCA):}
It has been shown that the performance of unified multimodal training may decline over that of individual modalities due to the differences in learning dynamics, noise topologies, etc. \cite{9156420}. Therefore, we have used fixed feature vectors of audio and visual modalities to train the proposed A-V fusion model. %We have explored a recursive fusion of the attended features obtained from the joint cross-attentional model to obtain more refined feature representations. % by effectively capturing the intra and inter-modal relationships across audio and visual modalities. 
By deploying the joint feature representation in the CA framework in a recursive fashion, we are able to simultaneously enhance the modeling of both intra- and inter-modal relationships among audio and visual modalities.
%intra- and inter-modeling of A-V relationships. 
The block diagram of the proposed approach is shown in Fig. \ref{Block Diagram}. The joint representation $\boldsymbol{J}$ is obtained by concatenating the audio and visual feature vectors as
%\begin{equation}
 ${\boldsymbol J} = [{\boldsymbol X}_{a} ; {\boldsymbol X}_{v}] \in\mathbb{R}^{d\times L}$,
%\end{equation}
where $d = {d_a} + {d_v}$ denotes the feature dimension of concatenated features. The concatenated A-V feature representations ($\boldsymbol J$) of the given video subsequence ($S$) are now used in the CA framework to attend to the individual modalities. This helps to incorporate both intra- and inter-modal relationships in obtaining the attention weights of audio and visual modalities. Now the correlation across joint feature representation and the individual modalities are obtained as joint cross-correlation matrix, which is given by  
\begin{align}
%\begin{equation}
   \boldsymbol C_{a}= \tanh \left(\frac{{\boldsymbol X}_{a}^\top{\boldsymbol W}_{ja}{\boldsymbol J}}{\sqrt d}\right);
%   \end{equation}
%\text{and} 
%where ${\boldsymbol W}_{\mathbf j \mathbf a} \in\mathbb{R}^{L\times L} $ represents learnable weight matrix across the audio and combined audio-visual features, and $T$ denotes transpose operation. Similarly, the joint correlation matrix for visual features is given by: 
%\begin{equation}
   \boldsymbol C_{v}= \tanh \left(\frac{{\boldsymbol X}_{v}^\top{\boldsymbol W}_{jv}{\boldsymbol J}}{\sqrt d}\right)
%\end{equation}
\end{align}
where ${\boldsymbol W}_{ja} \in\mathbb{R}^{{d_a}\times d}$, ${\boldsymbol W}_{jv} \in\mathbb{R}^{{d_v}\times d} $ represents learnable weight matrices of audio and visual modalities respectively. %, and $\top$ denotes transpose operation.

The joint correlation matrices $\boldsymbol C_{a}$ and $\boldsymbol C_{v}$ for audio and visual modalities help to obtain the attention weights based on the semantic relevance of both across and within the modalities. The higher the correlation coefficient, the higher the correlation across the corresponding samples within the same modality as well as across the modalities. Now the joint cross-correlation matrices are used to obtain the attention maps of audio and visual modalities, which are given by   
\begin{align}
%\begin{equation}
\boldsymbol H_{a}=ReLU(\boldsymbol X_{a} \boldsymbol W_{ca} {\boldsymbol C}_{a})\\
%\end{equation}
%\begin{equation}
\boldsymbol H_{v}=ReLU(\boldsymbol X_{v} \boldsymbol W_{cv} {\boldsymbol C}_{v})
%\end{equation}
\end{align}
where ${\boldsymbol W}_{ca} \in\mathbb{R}^{{L}\times {L}} $,${\boldsymbol W}_{cv} \in\mathbb{R}^{{L}\times {L}} $ denote learnable matrices of audio and visual modalities respectively. These attention maps are used to obtain the attended features of audio and visual modalities as follows:
\begin{align}
%\begin{equation}
{\boldsymbol X}_{att, a} = \boldsymbol H_{a} \boldsymbol W_{ha} + \boldsymbol X_{a}\\
%\end{equation}
%\begin{equation}
{\boldsymbol X}_{att, v} = \boldsymbol H_{v} \boldsymbol W_{hv} + \boldsymbol X_{v}  
%\end{equation}
\end{align}
where $\boldsymbol W_{ha} \in\mathbb{R}^{L\times {L}}$ and $\boldsymbol W_{hv} \in\mathbb{R}^{L\times {L}}$ denote the learnable weight matrices for audio and visual modalities respectively. To obtain more refined feature representations, the attended features are again fed as input to the joint cross-attentional model, which is given by 
\begin{align}
%\begin{equation}
{\boldsymbol X}_{att, a}^{(t)} = \boldsymbol H_{a}^{(t)} \boldsymbol W_{ha}^{(t)} + \boldsymbol X_{a}^{(t-1)}\\
%\end{equation}
%\begin{equation}
{\boldsymbol X}_{att, v}^{(t)} = \boldsymbol H_{v}^{(t)} \boldsymbol W_{hv}^{(t)} + \boldsymbol X_{v}^{(t-1)}  
%\end{equation}
\end{align}
where $\boldsymbol W_{ha}^{(t)} \in\mathbb{R}^{L\times {L}}$ and $\boldsymbol W_{hv}^{(t)} \in\mathbb{R}^{L\times {L}}$ denote the learnable weight matrices of audio and visual modalities respectively and $t$ refers to the recursive step. 

Finally the attended audio and visual features, ${\boldsymbol X}_{att,a}^{(t)}$ and $ {\boldsymbol X}_{att,v}^{(t)}$ obtained from the recursive fusion model are concatenated and fed to the BLSTMs followed by attentive statistics pooling (ASP) \cite{okabe18_interspeech} to obtain semantic utterance-level representation of the A-V feature vectors. In this work, BLSTMs are used to enhance the temporal modeling of A-V feature representations. %similar to that of \cite{}. 
The utterance-level A-V feature representations are used to obtain the scores, where additive angular margin softmax (AAMSoftmax) \cite{8953658} loss function is used to optimize the parameters of the proposed model. %and ASP module.   

%are further concatenated to obtain the A-V feature representation, which is given by:  
%\begin{equation}
%\mathbf {\widehat X} = [{\boldsymbol X}_{\mathbf a\mathbf t\mathbf t\boldsymbol,\mathbf v} ; {\boldsymbol X}_{\mathbf a\mathbf t\mathbf t\boldsymbol,\mathbf a} ]  %
%\end{equation}
%The attended audio-visual feature vectors are fed to the Bi-directional LSTM in order to capture the temporal dynamics of the attended joint audio-visual feature representations. The segment-level audio-visual feature representations are in turn fed to the attentive statistics pooling (ASP) \cite{okabe18_interspeech} in order to obtain the sub-sequence or utterance-level representation of the audio-visual feature vectors. 

\section{Results and Discussion}
%\subsection{Dataset and Evaluation Metrics}
\noindent \textbf{A) Dataset and Evaluation Metrics:}
%\textbf{Voxceleb1 Dataset:} 
We have evaluated the proposed model on the Voxceleb1 dataset \cite{Nagrani17}, obtained from YouTube videos under challenging environments. The dataset consists of 1,48,642 video clips, captured from 1251 speakers, of which 55\% of the speakers are male. Each video clip has a duration of 4 to 145 seconds, and the speakers are chosen to cover a diverse range of ethnicities, accents, professions, and ages. For our experiments, we have divided the voxceleb1 development set, which has 1211 speakers into training and validation sets. The training and validation splits were randomly selected as 1150 and 61 speakers, respectively, and our results are reported on both the validation split and Vox1-O (Voxceleb1 original) test set for performance evaluation. It is worth mentioning that the models are trained only on the Voxceleb1 dataset.    

The performance of the proposed approach is evaluated using the Equal Error Rate (EER) and minimum Detection Cost Function (minDCF), which has been widely used for speaker verification in the literature \cite{9922810,9889705}. Compared to EER, DCF offers control over the contribution of false positives and false negatives based on their prior probabilities \cite{BRUMMER2006230}. In our experiments, we considered the parameters of the DCF as $P_{target}=0.05$, $C_{miss}=1$ and $C_{falsealarm}=1$ similar to that of VoxSRC-20 \cite{nagrani2020voxsrc}.
\noindent \textbf{B) Ablation Study:}
The results are reported based on the average of three runs for statistical stability. To obtain the audio and visual feature vectors, we have used Resnet-18 \cite{7780459} for visual modality and ECAPA-TDNN \cite{Desplanques2020} for audio modality similar to that of \cite{10096814} in order to have a fair comparison. We have conducted a series of experiments to analyze the performance of the proposed approach by comparing it with some of the widely used fusion strategies shown in Table \ref{Block Diagram}. First, we have implemented a simple score-level fusion, where scores are obtained from individual modalities and then fused to verify the identity of a person. We also explored another widely used fusion strategy of feature concatenation, where the features of the audio and visual modalities are concatenated and used to obtain the final score. We can observe that the fusion performance has been improved over simple score-level fusion. To analyze the impact of the proposed approach, we further compared it with some of the widely explored attention mechanisms. By employing a self-attention mechanism on the concatenated features of individual modalities, the fusion performance has been further improved by leveraging the intra-modal relationships. We also explored inter-modal relationships across the modalities using CA and found further improvement in fusion performance over prior fusion strategies. Now, we employed joint cross-attention, where joint feature representation is integrated into the CA framework to simultaneously capture both intra- and inter-modal relationships. Finally, we have introduced the proposed recursive fusion with joint CA to obtain the attended features of individual modalities, and we can observe that recursive fusion helps in obtaining more refined feature representations and achieves the best performance among all the fusion strategies.             

To further analyze the contribution of the individual components of the proposed approach, we conducted another series of experiments with BLSTMs and varying the number of iterations of the recursive mechanism. First, we performed two experiments without recursion, one with BLSTM and another without BLSTM to understand the contribution of BLSTM to the proposed approach. Table \ref{tab2} shows the results of the experiments on the validation set of the Voxceleb1 dataset. We can observe that BLSTMs improve the fusion performance by taking advantage of the temporal modeling of the A-V feature representations. To understand the impact of the recursive fusion, we performed experiments by varying the number of iterations and obtained the best performance at 3 iterations. Beyond that, we observe a decline in the fusion performance, which can be attributed to the fact that though recursion helps in improving fusion performance initially, it results in overfitting with more iterations, resulting in a performance decline on the validation set.
%acts as a regularizer and improves the generalization ability of the proposed model. 
%We also performed experiments with multiple iterations but without BLSTMs and found a similar trend of improvement in fusion performance. % and achieved the best result with 3 iterations.          
\begin{table}
\centering
\caption{Performance of various fusion strategies on the validation set}\label{tab1}
\begin{tabular}{|c|c|c|}
\hline
\textbf{Fusion} &   \multicolumn{2}{|c|}{\textbf{Validation Set}} \\
\cline{2-3}
\textbf{Method} &   \textbf{EER $\downarrow$} &  \textbf{minDCF $\downarrow$} \\
\hline
\hline
Score-level Fusion &  2.521 & 0.217 \\ \hline
Feature Concatenation  & 2.489 & 0.193 \\ \hline
Self-Attention &  2.412 & 0.176\\ \hline
Cross-Attention &  2.387 & 0.149 \\ \hline
Joint Cross-Attention &  2.315 & 0.135\\ \hline
RJCA (w/o BLSTMs) &  1.946 & 0.128 \\ 
\hline
RJCA (w/ BLSTMs) &  \textbf{1.851} & \textbf{0.112} \\ 
\hline
\end{tabular}
\end{table}
\begin{table}
\centering
\caption{Performance of proposed model with varying number of iterations and BLSTM on the validation set}\label{tab2}
\begin{tabular}{|c|c|c|}
\hline
\textbf{Fusion} &  \multicolumn{2}{|c|}{\textbf{Validation Set}} \\
\cline{2-3}
\textbf{Method} &   \textbf{EER $\downarrow$} &  \textbf{minDCF $\downarrow$} \\
\hline
\multicolumn{3}{|c|}{\textbf{RJCA Fusion w/o Recursion}}
\\ \hline
RJCA Fusion w/o BLSTMs & 2.315 & 0.135\\ \hline
RJCA Fusion w/ BLSTMs & 2.195 & 0.132 \\ \hline
\multicolumn{3}{|c|}{\textbf{RJCA Fusion w/ Recursion}} \\ \hline
RJCA Fusion w/o BLSTMs, t = 3 & 1.946 &   0.128\\ \hline % Best performance w/o transformers
RJCA Fusion w/ BLSTMs, t = 2 & 2.029 & 0.119\\ \hline
RJCA Fusion w/ BLSTMs, t = 3 & \textbf{1.851} & \textbf{0.112} \\ \hline
RJCA Fusion w/ BLSTMs, t = 4 & 1.982 & 0.124\\ \hline
RJCA Fusion w/ BLSTMs, t = 5 & 2.159 & 0.137 \\ \hline
\end{tabular}
\end{table}
\begin{table}
\centering
\caption{Performance of the proposed approach in comparison to state-of-the-art on the validation set and Vox1-O set }\label{tab3}
\begin{tabular}{|c|c|c|c|c|c|c|c|c|c|c|}
\hline
\textbf{Fusion} &   \multicolumn{2}{|c|}{\textbf{Validation Set}} & \multicolumn{2}{|c|}{\textbf{Vox1-O Set}} \\
\cline{2-5}
\textbf{Method} &  \textbf{EER $\downarrow$} &  \textbf{minDCF $\downarrow$} & \textbf{EER $\downarrow$} &  \textbf{minDCF $\downarrow$} \\
\hline
\hline
Visual & 3.720 & 0.298 & 3.779 & 0.274 \\ \hline
Audio & 2.553 & 0.253 & 2.529  & 0.228 \\ \hline
Tao et al \cite{10096814} &  2.476 &  0.203 & 2.409 & 0.198\\ \hline
Sari et al \cite{9414260} &  2.438  & 0.178 & 2.397  & 0.169 \\ \hline
Chen et al \cite{Chen2020} & 2.403  & 0.163 & 2.376  & 0.161 \\ \hline
Mocanu et al \cite{9922810} & 2.387  & 0.149 & 2.355 & 0.156 \\ \hline
Praveen et al \cite{praveen2023audiovisual} & 2.173  & 0.126 & 2.214 & 0.129 \\ \hline
RJCA (w/ BLSTMs) & \textbf{1.851} & \textbf{0.112} & \textbf{1.975} & \textbf{0.116} \\
\hline
\end{tabular}
\end{table}
%\subsection{Comparison to state-of-the-art}

\noindent \textbf{C) Comparison to state-of-the-art:}
%Most of the existing methods used the Voxceleb2 development dataset for training the models for person verification. However, 
We have used the Voxceleb1 dataset to validate the proposed approach and compared it with the relevant state-of-the-art methods using A-V fusion for person verification. %We have used the same experimental setup as that of \cite{10096814} in order to have a fair comparison. 
To have a fair comparison, we have reimplemented the state-of-the-art methods using our experimental setup due to the differences in their experimental protocols. Table \ref{tab3} shows the comparison of the proposed approach with state-of-the-art methods and individual modalities on both the validation split of the Voxceleb1 and Vox1-O partitions. First, we have conducted experiments with the individual modalities and found that audio performs relatively better than visual modality. %In order to have a fair comparison, we have re-implemented the approach of \cite{10096814} using the same experimental setup and training on the Voxceleb1 dataset. 
%Sari et al \cite{9414260} explored a shared representation for audio and visual modalities using a shared classifier, 

Tao et al. \cite{10096814} explored complementary relationships between the audio and visual modalities as supervisory information to mitigate the impact of noisy labels. We have used the clean samples obtained using \cite{10096814} to train our models as well as other methods \cite{9414260,Chen2020,9922810,praveen2023audiovisual}. Most of the existing approaches \cite{9414260,Chen2020,10096814} rely on a single image (randomly chosen), thereby focusing on the inter-modal relationships, failing to retain the intra-modal relationships. Both of these approaches \cite{9414260} and \cite{Chen2020} are able to improve the fusion performance better than that of \cite{10096814}, where \cite{9414260} uses a shared representation space using a shared classifier and \cite{Chen2020} explored the gating mechanism. Mocanu et al. \cite{9922810} also focused on capturing the inter-modal relationships based on cross-correlation across the modalities. %where NetVLAD \cite{8683120} is used for aggregating the audio and visual features. 
Unlike these approaches, Praveen et al. \cite{praveen2023audiovisual} showed improvement by capturing both intra- and inter-modal relationships simultaneously using joint feature representation in the CA framework. We have further improved the fusion performance by introducing recursive fusion to obtain more refined feature representations.   
%Sari et al \cite{9414260} improved the performance using a shared representation space for audio and visual modalities using a shared classifier. 
%Mocanu et al \cite{9922810} explored cross-attention based on cross-correlation across the modalities and showed improvement in performance better than that of \cite{10096814}. 
Since the proposed RJCA model leverages both intra- and inter-modal relationships effectively, we can observe that the fusion performance has been improved over other state-of-the-art methods. %Since the proposed approach introduces the recursive fusion along with BLSTMs to further enhance the A-V feature representations, the fusion performance has been further improved than that of \cite{10096814}. % and \cite{praveen2023audiovisual}.   

%%%%%%%%%%%%%%%%%%%%%%%%%%%%%%%%%%%%%%%%%%%%%%%%%%%%%%%%%%%%%%%%%%%%%%%%%%%%%%%%
\section{CONCLUSION}
In this paper, we have presented a novel approach of recursive joint CA fusion for person verification by effectively exploiting both inter- and intra-modal relationships across audio and visual modalities. Specifically, we have explored the joint feature representation in the CA framework recursively to obtain more refined A-V feature representations. To further enhance the intra-modal relationships of A-V feature representations, we have used BLSTMs on the refined A-V feature representations obtained from the proposed RJCA model. We showed that the fusion performance can be improved by effectively capturing the intra- and inter-modal relationships. The performance of the proposed approach can be further enhanced by training with the large-scale Voxceleb2 dataset as it can improve the generalization ability of the proposed approach.      

%\section{ACKNOWLEDGEMENT}
%The authors wish to acknowledge the funding from the Government of Canada’s New Frontiers in Research Fund (NFRF) through grant NFRFR-2021-00338.
% Authors acknowledge the funding from NFRF through NFRFR-2021-00338
%%%%%%%%%%%%%%%%%%%%%%%%%%%%%%%%%%%%%%%%%%%%%%%%%%%%%%%%%%%%%%%%%%%%%%%%%%%%%%%%
%\section{ACKNOWLEDGMENTS}

%The authors gratefully acknowledge the contribution of reviewers' comments, etc. (if desired). Put sponsor acknowledgments in the unnumbered footnote on the first page.

%%%%%%%%%%%%%%%%%%%%%%%%%%%%%%%%%%%%%%%%%%%%%%%%%%%%%%%%%%%%%%%%%%%%%%%%%%%%%%%%

{\small
\bibliographystyle{ieeetr}
\bibliography{egbib}
}

\end{document}